\documentclass{article}
\usepackage{graphicx}
\usepackage{etoolbox}
\AtBeginEnvironment{quote}{\par\small}

\usepackage[backend=biber,natbib=true,style=numeric]{biblatex}

\bibliography{ELIZA_Reanimated.bib}
\title{ELIZA Reanimated: The world's first chatbot restored on the world's first time sharing system}
\usepackage{authblk}
\author[1]{Rupert Lane}
\affil[1]{rupert@timereshared.com}
\author[2]{Anthony Hay}
\affil[2]{anthony.hay.1@gmail.com}
\author[3]{Arthur Schwarz}
\affil[3]{home@slipbits.com}
\author[4]{David M. Berry}
\affil[4]{D.M.Berry@sussex.ac.uk}
\author[5]{Jeff Shrager}
\affil[5]{JShrager@Stanford.edu}
\date{January 2025}

\begin{document}

\maketitle

\begin{abstract}

ELIZA, created by Joseph Weizenbaum at MIT in the early 1960s, is usually considered the world's first chatbot. It was developed in MAD-SLIP on MIT's CTSS, the world's first time-sharing system, on an IBM 7094. We discovered an original ELIZA printout in Prof. Weizenbaum's archives at MIT, including an early version of the famous DOCTOR script, a nearly complete version of the MAD-SLIP code, and various support functions in MAD and FAP. Here we describe the reanimation of this original ELIZA on a restored CTSS, itself running on an emulated IBM 7094. The entire stack is open source, so that any user of a unix-like OS can run the world's first chatbot on the world's first time-sharing system. 

\end{abstract}

``We can only see a short distance ahead, but we can see plenty there that needs to be done.'' (The last line of Turing's 1950 MIND paper \cite{AT50Mind})

\section{ELIZA's Ancestry}

If Alan Turing was AI's founding father, Ada Lovelace may well have been its founding mother. Over a century before Turning famously proposed using the Imitation Game to determine whether a computer is intelligent \cite{AT50Mind}, Lady Lovelace described the potential of Charles Babbage's Analytical Engine to ``act upon other things besides number, were objects found whose mutual fundamental relations could be expressed by those of the abstract science of operations, and which should be also susceptible of adaptations to the action of the operating notation and mechanism of the engine.''\cite{ADA43} She gives the example of music: ``Supposing, for instance, that the fundamental relations of pitched sounds in the science of harmony and of musical composition were susceptible of such expression and adaptations, the engine might compose elaborate and scientific pieces of music of any degree of complexity or extent.'' 

Ada's prescient insight that machines could act upon entities besides numbers foreshadowed symbolic computing which, in the 1950s, a mere moment after Turing's famous paper, arose, and remains today, one of the foundations of artificial intelligence \cite{NEWELLSIMON81}. Symbolic computing, along with list processing, has seen application in every domain, particularly in Natural Language Processing (NLP), the linage that begat chatbots, of which Joseph Weizenbaum's ELIZA is considered the earliest exemplar \cite{JW66ELIZA}.\footnote{Of course, ELIZA would not have been called a ``chatbot'' at that that time as that term was not invented until the mid 1990's \cite{chatbot}} 

But before ELIZA, Weizenbaum was known as the inventor of SLIP, the ``Symmetric Lisp Processor''\cite{SLIP}, a significant contender in the line of ``list-processing'' languages beginning with IPL-V, developed at RAND to build the world's first AIs \cite{IPLManual}\cite{LT}, and including Lisp \cite{HistoryofLisp}. IPL was the direct ancestor of SLIP and LISP, introducing the core computer science methods of list processing, symbolic computing, and recursion. But it was a very unwieldy language, being essentially the machine language for an emulated stack machine\footnote{IPL's built in functions were named like ``J123''!} and, as an interpreted language it was slower than assembly or the soon-to-follow ``high-level'' compiled languages that promised programmers both good performance and simplicity of expression. There were several projects mounted to build on the successes of Newell and Simon's IPL work without having to cope with its ugliness and inefficiency. The high-level languages, notably Fortran and COBOL were aimed at engineering and business applications, and did not provide the AI-related functionalities of IPL, such as symbol processing, lists, and recursion. So the question naturally arose as to how to add these capabilities to those already-existing languages. 

Weizenbaum created SLIP specifically to fill this void, originally as a library to be used with Fortran. On the strength of this work, Weizenbaum was invited to join Project MAC at MIT \cite{ProjectMAC}\footnote{Project MAC was founded, in part, by John McCarthy, the inventor of Lisp and the person who coined the term ``Artificial Intelligence''\cite{WPDartAIConf}, although McCarthy moved to Stanford in 1962, two years before Weizenbaum arrived at MIT \cite{WPJM}.} One of the core goals of Project MAC was to demonstrate the power of the new concept of interactive time-sharing, embodied by CTSS \cite{CTSS}, the ``Compatible Time-Sharing System'', the world's first interactive time-sharing system. CTSS was originally conceived and implemented by MIT Computation Center personnel, led by Fernando Corbató. Project MAC improved CTSS and made it available to many different projects, including Weizenbaum's.

CTSS's core user-level language was MAD \cite{MADManual}, so Weizenbaum rebuilt SLIP for MAD, and shortly thereafter built ELIZA in MAD-SLIP on CTSS. In addition to being an excellent demonstration of the power of interactive time-sharing, ELIZA was intended as an exploration in symbolic computing applied to NLP, and as a platform for mounting experiments in human-machine interaction around discourse \cite{JSArxiv}. Although it is not clear whether ELIZA was intended as an AI or, to the contrary, as a bit of an anti-AI, Weizenbaum was closely connected to AI through a number of paths, including Kenneth Colby, a Stanford psychiatrist who was interested in modeling neurosis and paranoia \cite{C&G64}, and Ed Feigenbaum, who had been a student of Simon's and was creating various AI and cognitive simulation programs in IPL \cite{Feigenbaum1959}. And intentionally or not, ELIZA has had a significant impact on AI, being the first program to directly embody Turing's test, and to bring to reality the science-fiction fantasy of holding an actual conversation with a computer.

\section{ELIZA's Descendancy}

Weizenbaum foresaw that ``[t]here is a danger [...] that the example [ELIZA] will run away with what it is supposed to illustrate''\cite[43]{JW66ELIZA}. Indeed, it did! Through interesting and complex criss-crossing histories, described in detail in \cite{BBNHistory}, Danny Bobrow, a recent MIT AI graduate who headed up BBN's AI program, brought McCarthy, and thus Lisp, to BBN, where Bernie Cosell was exposed to it. Shortly after the MAD-SLIP ELIZA appeared in CACM, Cosell wrote a near-clone of it in Lisp \cite{PS09CAW}.\footnote{Bobrow was also the author of STUDENT, which solved algebra word problems stated in natural language \cite{bobrowStudent}.}  

Shortly after Cosell wrote his Lisp ELIZA, BBN became the leading technical contractor for the nascent ARPAnet, which of course begat the Internet and thence The Web. BBN built the ARPAnet's core hardware and software \cite{WPARPANET}. Once the ARPANet was up, with BBN as one of the founding sites, Cosell's Lisp ELIZA diffused rapidly through that network and across the soon-to-be Lisp-centered world of academic AI. As a result, Cosell's Lisp ELIZA rapidly became the dominant strain, and Weizenbaum's MAD-SLIP version, invisible to the ARPAnet, was left to history.\footnote{the Project MAC 7094 that ran CTSS, and thus the original ELIZA, was never on the ARPAnet, although other Project MAC computers were. DEC PDP-10s, which were available in both Project MAC and at the AI Lab, also located at Tech Square, were among the first academic machines to join the ARPANet, the PDP-10 being a common academic machine at the time. Therefore, even as Joseph Weizenbaum's MAD-SLIP ELIZA was running on the isolated CTSS machines, Bernie Cosell's Lisp ELIZA was running just a few floors away on the ARPAnet-linked computers at the AI Lab, and probably even just a few steps away on ARPANet-linked machines at Project MAC as well. As a result, numerous MIT AI students who would go on to be the seed AI professors of the latter 20th century were exposed to the Lisp ELIZA, not the original in MAD-SLIP \cite{SmoliarPersonal}.}

Cosell reports: 

\begin{quote}
``When I was working on the PDP-1 time-sharing system [...] I thought I would learn Lisp. That spring, Joe Weizenbaum had written an article for Communications of the ACM on ELIZA. I thought that was way cool. [...] He described how ELIZA works and I said, `I bet I could write something to do that.' And so I started writing a Lisp program on [the] PDP-1 system at BBN.''\cite[540]{PS09CAW} He continues, ``I wrote that program and got it up and working. Playing with it was an all-BBN project. [...] It was written, at first, in the PDP-1 Lisp. But they were building a Lisp on the PDP-6 at that point—or maybe the PDP-10. But it was the Lisp that had spread across the ARPANet. So [ELIZA] went along with it [...].''\cite[541]{PS09CAW} 
\end{quote}

Once Cosell's Lisp ELIZA hit the academic world via the rapid spread of the ARPANet, Weizenbaum's MAD-SLIP version was no longer relevant, and the name ``ELIZA'' (and the ``DOCTOR'' script), as well as the concept, was, from that point forward, associated with Cosell's Lisp version, although its origin was still correctly attributed to Weizenbaum via the CACM paper, leading to a 50-year-long community-wide misapprehension that ELIZA had been written in Lisp.\footnote{To make matters even worse for the original MAD-SLIP ELIZA, the DOCTOR script, \textit{but not the MAD-SLIP code}, was published in its entirety (although with errors, as described in section \ref{scriptediting}) in Weizenbaum's 1966 paper. Both Lisp and SLIP used identical representation (called s-expressions in Lisp) to represent lists, and the DOCTOR script, being a series of lists, was represented using this syntax in the CACM paper. Anyone merely skimming that paper would not have seen any MAD-SLIP code, but would have seen the complete DOCTOR script which \textit{appeared} to be a Lisp s-expression.} In addition to being promulgated by Cosell's Lisp clone being the one that was most easily available via the ARPANet, Lisp was rapidly becoming the go-to language of AI.

Another defining event in the descendancy of the original ELIZA occurred almost exactly a decade later, in 1977. Creative Computing, one of the magazines that served as the ``GitHub'' of the mid-70s personal computer explosion, published an ELIZA knock-off written in BASIC \cite{SNCC77}, coincidentally coinciding with the so-called ``1977 trinity'' -- the year that the Commodore Pet, the Apple II, and the TRS-80 all appeared \cite{WPHPC} . Within a few years millions of computer hobbyists had personal computers of all sorts, mostly with BASIC as their primary user-level programming language, and probably not a small number of those hobbyists were interested enough by the possibility of AI to type in this BASIC ELIZA (which was only a couple of pages of code), and experiment with it themselves. Because of its brevity and simplicity, and the personal computer explosion, this ELIZA begat hundreds of knock-offs through the decades, in every conceivable programming language, making it perhaps the most knocked-off program in history.\footnote{Jeff Shrager (JS) curates a web site, ELIZAGen.org \cite{Elizagen}, dedicated to the history of ELIZA and ELIZA-like programs. In that capacity he is regularly sent new, or newly-discovered knock-offs of one or another of the ELIZA threads, usually these are knock-offs of his own BASIC ELIZA. In fact, in a coincidence too complex to untangle, Shrager's BASIC ELIZA was re-translated into Lisp for the APPLE-][ \cite{plisp}, and published, without his being aware of it, as an appendix for the P-Lisp user's manual, a manual that \textit{Shrager himself had co-authored.}} Just as Cosell's Lisp ELIZA spread via the ARPANet, the BASIC ELIZA spread via the explosive contagion of personal computers. 

As a result of these coincidences and an inherent interest in AI (or at least in talking with computers), the version of ELIZA that was known in the academic community was usually written in Lisp (starting with Cosell's clone), and the version known to the public was usually written in BASIC (starting with Shrager's knock-off). But until it was rediscovered in 2021, the original MAD-SLIP ELIZA had not been seen by anyone for at least 50 years \cite{findingELIZA,Elizagen}.

\section{ELIZA Rediscovered}

MAD was soon replaced by other ALGOL-like languages, and Lisp's widespread adoption eclipsed SLIP. Few aside from a tiny cadre of AI historians concerned themselves with the whereabouts of the original ELIZA code until 2021 when Jeff Shrager -- the author of the 1973/1977 BASIC ELIZA -- and MIT archivist Myles Crowley found what appeared to be a complete copy of the source among Weizenbaum's papers \cite{ELIZASource}. Shrager contacted the Weizenbaum estate who granted permission to open source the code \cite{oseliza}. In a \textbf{CoRecursive} podcast of July 2022 Shrager recounts the moment of discovery:

\begin{quote}
    ``[Myles and I] pulled the box [labeled ...] `computer conversations, box eight'. [I said ...] we’re looking for some code [...]. We opened [...] Folder one, box eight, and [...] there’s ELIZA, there’s [some version printed out of] the original ELIZA with the [...] almost exact doctor script. [...] We [also] found many conversations with ELIZA that nobody had ever seen before, many of them hand edited by Weizenbaum.''\cite{podcast} 
\end{quote}

Shrager (at that time, and in a later exploration of the archives undertaken by Shrager and David Berry) also found relevant underlying code, including critical parts of MAD-SLIP, written in both MAD and FAP, the 'Fortran Assembler Language'.\footnote{Its name notwithstanding, FAP was used to add machine-level code to MAD as well as Fortran. Fortran was available on the 709x in batch environments before CTSS. However, the Fortran compiler was too heavy to adapt to be used interactively, so the much smaller MAD was CTSS's main high level language. FAP code was required to call CTSS services from MAD, and was also used to carry out various functions that were easier to express at machine-level, such as bit-wise and word-wise manipulations.}    

\section{ELIZA Analyzed}\label{missingscriptfunctions}

Simply reading the MAD-SLIP code revealed that there were several functionalities described in the 1966 paper that were not in the found version:\footnote{The reader can refer to the Weizenbaum's 1966 CACM paper  \cite{JW66ELIZA} for explanations of these, although it is somewhat obvious from the context what they are intended to do.}

\begin{itemize}
\item The preliminary transformation \texttt{PRE} reassembly pattern is not implemented, for example:
\begin{verbatim}
(I'M = YOU'RE ((0 YOU'RE 0) 
   (PRE (YOU ARE 3) (=I))))
\end{verbatim}

\item Neither the keyword stack nor the \texttt{NEWKEY} reassembly rule are supported, for example:
\begin{verbatim}
(DREAMT 4 ((0 YOU DREAMT 0)
   (REALLY, 4) 
   (HAVE YOU EVER FANTASIED 4 WHILE YOU WERE AWAKE)
   (HAVE YOU DREAMT 4 BEFORE)
   (=DREAM)
   (NEWKEY)))
\end{verbatim}

\item The found source code supports links at the transformation rule level, for example: 
\begin{verbatim}
(HOW (=WHAT))
\end{verbatim}
But does not support them at the reassembly level, as in the \texttt{(=DREAM)} as used in the \texttt{DREAMT} rule above.
\end{itemize}

As was mentioned, the version discovered by Shrager came physically attached (on the fan-fold printout) to a simpler version of the DOCTOR script than that in the 1966 paper. The found code also appeared to have an internal editing capability that was only mentioned in passing in a single sentence in the 1966 CACM paper, and is described in more detail in section \ref{teach}.

Although it was clear from these omissions that the found version was not exactly the same as the published ELIZA what remained unknown was, aside from the obvious missing functionality, whether this version of ELIZA worked at all; For all we knew we might have found the printout of a version riddled with bugs. The best way -- indeed perhaps the only way -- to figure out if this was a working ELIZA was to try it. 

\section{ELIZA Reanimated}

Reanimating ELIZA from the discovered code may have been the best route to proving that we had a ``real'' ELIZA in our hands, but it was not simple! It required numerous steps of code cleaning and completion, emulator stack installation and debugging, non-trivial debugging of the found code itself, and even writing some entirely new functions that were not found in the archives or in the available MAD and SLIP implementations.  

Rupert Lane (RL), an aficionado of early operating systems, had been working with MTS (The Michigan Time Sharing system, which has MAD and SLIP implementations). David Berry (DB) ask him to think about bringing up the original ELIZA. After some discussion with the rest of the team we realized that it would be simpler to bring up the original ELIZA in its original environment of CTSS on the 7094.

\subsection{CTSS on the 7094}

MIT's CTSS, first introduced in 1963 at MIT, was the first multi-user time sharing system \cite{CTSS}. It brought many innovations such as a disk filing system, real time communication with users via typewriter terminals, and the ability to develop programs interactively without waiting hours or days for batch execution. It would have been impossible for Weizenbaum to develop ELIZA without such a system.

CTSS ran on the IBM 7094, an early transistorized computer with only 32k of user memory, a 36 bit word length, and a clock speed of around 450kHz. At that time a 7094 cost \$2.9 million (equivalent to \$23 million in 2023), and MIT heavily modified it, including adding a second 32k memory bank for system software. CTSS could support around 30 simultaneous users and was in use at MIT until the early 1970s.

An emulator for the 7094 was written by David Pitts, based on work by Paul Pierce, and a working version of CTSS was brought up on that machine by David in the early 2000s \cite{CTSSemulator}. 

\subsection{Restoring the Code}

The combination of MAD and SLIP that constitutes the ELIZA stack is around 2600 lines of mostly un-commented MAD and FAP code. Getting this into machine readable format was non-trivial; it did not OCR well\footnote{OCR is optimized for modern fonts and natural language, not 60 year old programs with extremely obscure abbreviated operator names printed in line-printer fonts with fading ink.}, so Anthony Hay (AH) and Art Scwartz (AS) manually transcribed most of it. AH had previously done a careful analysis of the algorithm and script as described in the original 1966 paper, and wrote a C++ version of ELIZA that is, so far as we know, the best clone of the original \cite{anteliza}. AS has worked with SLIP for many years and is the author of the GNU SLIP library \cite{ArtGNUSLIP}. Their expertise was invaluable in navigating the code.

RL fed each file into the compiler on the virtual CTSS system, which helped hunt down typos and let us understand the structure of the code. Some of the unique challenges of working with this code included:
\begin{itemize}
    \item CTSS uses 6 bit BCD character encoding, packing six characters into a 36 bit word. (This was before bytes or ASCII were invented!)
    \item The MAD language uses long keywords like WHENEVER, but allows the programmer to abbreviate them to things like W'R. Although understandable in the era of punch cards, this makes the code difficult to read (and OCR).
    \item Source code was often created on punch cards at that time, so column layout is important. For example, to enter lines longer than 78 characters, one punched a 1 in column 11 of the continuation card. 
\end{itemize}

\subsubsection{Missing Functions and other Minor Issues}
As we got each piece compiled, we realized that several functions were missing:
\begin{itemize}

\item  \underline{\texttt{BCDIT}}, used in function \texttt{FRBCD}, appears to be converting a binary number (the rightmost 18 bits according to the multiplication of K) to a string of BCD. We replaced this with the CTSS library routine \texttt{DELBC}.

\item \underline{\texttt{INLST}} is used in \texttt{XMATCH} and \texttt{ASSMBL}, two of the most critical ELIZA functions. It appears in the Fortran implementations of SLIP, and seems to take the cells of the first parameter and add them to the left of the list cells of the second parameter, so we wrote our own in MAD.

\item The \underline{\texttt{LETTER}} function was not in either the MAD or SLIP documentation, was not in the archive code, and google was not helpful. Inference from the calling code suggested that it classified characters in a word into one of 14 categories (An equal sign (`=') returns 1, a comma (`,') return 13, the digits `0'-`9' return 12, etc.) We wrote this function ourselves, in MAD. 

\item We had FAP code for \underline{\texttt{KGETBL}} that returned the desired entry in the right hand 6 bits but padded the remaining positions with spaces, so we masked these off. 

\item The MAD compiler on CTSS seems not to support \texttt{\$\$} in string constants to mean emit a single \texttt{\$}. The 1963 MAD manual confirms this, saying there is no clean way to put a \texttt{\$} in a string \cite[20]{MADManual}, but suggests this somewhat bizarre workaround: \texttt{TEST=-\$=\$} will set \texttt{TEST} to ``\texttt{\$     }''

\item Some initialization of common data areas appeared to be missing, causing the loader to fail.  We added code to 
ELIZA.MAD to initialize the public list \texttt{W(1)}: \texttt{LIST.(W(1))}\footnote{Interestingly, this appears to be directly borrowed from IPL, which also uses W1 (through 10) as temporary variables.}

\item The CTSS loader then failed due to too many object files CTSS did have a 'huge loader' where these limits are relaxed, but this component was not present in the emulated system. Luckily the source code for the loader was available in the operating system source tape dumps, so we compiled and added this.

\item With all the above changes, ELIZA compiled and completed reading the script for the first
time, and even began to interact for the first time in over 60 years. Unfortunately, it didn't get very far:

\begin{verbatim}
        r eliza
        W 850.1
        EXECUTION.
        WHICH SCRIPT DO YOU WISH TO PLAY
        100                             
         DOCTOR PLEASE TYPEWR PROBLE
        INPUT                       
        Men are all alike.

         PROTECTION MODE VIOLATION AT 23174.
         INS.=060671060671, RI.=000000000000, PI.=062061000000
        R .666+.000
\end{verbatim}

It took a fair amount of print-log debugging to narrow this down, but we finally found a single character error deep in the FAP code, at line 1670 of the function \texttt{SQIN.FAP}. We changed: ``\texttt{LAS     =H 0000}'' to: ``\texttt{LAS     =H 00000}''. (Of course it always comes down to a one charater typo!) 

\end{itemize}

\section{It's Alive!}

RL built a test environment where changes could be rapidly tested. We were eventually able to produce an executable, and on Saturday, December 21st, 2024 at 10:54(PDT), Rupert announced to the rest of the team:

\begin{verbatim}
I'm happy to announce that ELIZA is now running again on CTSS!

r eliza
        W 1835.0
        EXECUTION.
        WHICH SCRIPT DO YOU WISH TO PLAY
        200                             
        HOW DO YOU DO . PLEASE TELL ME YOUR PROBLEM
        INPUT                                     
        Men are all alike.
        IN WHAT WAY
\end{verbatim}

After all this effort, the reanimated ELIZA, not having run for over 60 years was able to carry on complete conversations.

\section{Doctor Under the Knife}\label{scriptediting}

There was still one problem standing between us and the goal of demonstrating that ELIZA actually worked. ELIZA uses a script containing the language rules that determine its personality. The most famous of these is, of course, the ``DOCTOR'' script which leads ELIZA to interact what seems to be a caricature of a Rogerian therapist. Weizenbaum chose this persona because he needed a context where ignorance would not destroy the illusion of understanding (``For example, in the psychiatric interview the psychiatrist says, tell me about the fishing fleet in San Francisco. One doesn't say, ``Look, he's a smart man—how come he doesn't know about the fishing fleet in San Francisco?'' What he really wants to hear is what the patient has to say about it.''\cite[251-253]{PM79MWT})

The code that we discovered in Weizenbaum's archives -- the code that we are running here -- had two early versions of the Doctor script attached (literally, in the sense of being on the same fan-fold print out). Their names are given in the printout as \texttt{.TAPE.100} and \texttt{.TAPE.102.} However, we had no example of a conversation with these scripts. Instead, we used a corrected version of the script published in the 1966 paper, which we called 200, because we could compare the I/O with the transcript published in the paper. Of course, we knew that some functionalities (e.g., \texttt{PRE}) are unimplemented in the present code. Regardless, in order to compare the found ELIZA's performance with the only published conversation, the one from the 1966 CACM paper, we used the script 200, the one Weizenbaum provided in that paper, with some corrections for typos that appeared in the original publication. In this way we were able to nearly exactly recreate the famous “Men are all alike” conversation that appeared in the 1966 paper. 

For his prior explorations in ELIZA, AH had a clean transcript of the DOCTOR script from the 1966 paper wherein he had already corrected several apparent typos, including lines that appear to have been replicated at each point in the fan-fold printout (every 34 lines). A few lines that were longer than 78 characters were reflowed.

We made one change in the prompt:
\begin{verbatim}
        He says I am depressed much of the time.
\end{verbatim}
The original prompt:
\begin{verbatim}
        He says I'm depressed much of the time.
\end{verbatim}
which,as a consequence of assuming that the \texttt{PRE} rule worked, produced:
\begin{verbatim}
        PRE 5:$5:$ 5QZ5QZ
\end{verbatim}
and then ELIZA might crash a few turns later. 

The other transcript difference is for the prompt:
\begin{verbatim}
        You are like my father in some ways.
\end{verbatim}
which always outputs: \texttt{= DIT} because this version of ELIZA does not understand \texttt{PRE} rule, and only supports \texttt{=keyword} references at the transformation rule as described in section \ref{missingscriptfunctions}, above.

Finally, using the 1966 script, the reanimated ELIZA followed the CACM dialogue except in those two places!
Here is a video of the reanimated ELIZA holding the conversation from the 1966 CACM paper: \cite{DemoVid}.

\subsection{Lady Ada's Revenge}

Running this early ELIZA proved that it in fact worked, and that it is very close to the ELIZA published in the 1966 CACM paper! However, running the actual program also led us to discover a significant bug that would have been very difficult to find by mere inspection of the code. To wit, this version of ELIZA does not handle numerical inputs correctly. More specifically, it will crash if given numeric input such as ``you are 999 today''. One could say that this version of ELIZA had taken Ada Lovelace's brilliant insight that computers could ``act upon other things besides number[s]'' a bit too literally. 

We tracked this problem to a complex issue in SLIP and the way it was being used by ELIZA. Without getting too into the weeds, when a number is entered it is turned into what SLIP interprets it a pointer which thence de-references to some arbitrary location in memory, generally leading to no good. We return to this issue in the section \ref{conc}.

\section{The Rain in Spain} \label{teach}

In his 1966 paper, Joseph Weizenbaum wrote mysteriously about being able to ``teach'' ELIZA: 
\begin{quote}
``Its name was chosen to emphasize that it may be incrementally improved by its users, since its language abilities may be
continually improved by a ``teacher''. Like the Eliza of Pygmalion fame, it can be made to appear even more civilized[...].''\cite{JW66ELIZA}
\end{quote}
We say ``mysteriously'' because one would think that, being the raison d'etre of the program's name, there would be significant discussion of this topic in the paper. However, this is the only mention of this capability anywhere in Weizenbaum's writings. Interestingly, the code we discovered and reanimated contained the complete functionality for this ``teacher'' mode, including functionality for editing and saving the script.

Typing `+' to ELIZA takes it out of conversation mode and into ``teaching'' mode. ELIZA responds with PLEASE INSTRUCT ME and waits for the user to enter any of seven commands: ADD \textless keyword\textgreater (\textless transformation rule\textgreater) adds the given transformation rule to an existing keyword rule; APPEND \textless keyword\textgreater (\textless existing reassembly rule\textgreater) (\textless new reassembly rule\textgreater) appends the new reassembly rule to the transformation rule with the existing reassembly rule in the given keyword; SUBST \textless keyword\textgreater (\textless existing reassembly rule\textgreater) (\textless new reassembly rule\textgreater) replaces the existing reassembly rule with the new one in the given keyword; TYPE \textless keyword\textgreater displays the script rule for the given keyword; DISPLA displays the rules for all keywords as well as the MEMORY rules; RANK \textless keyword\textgreater \textless n\textgreater allows the user to set the precedence for a keyword; and START returns ELIZA to the conversation mode. Furthermore, a special ``star'' (`*') command allows the teacher to add a rule more succinctly, without having to go into teaching mode. 

The Appendix contains a complete teaching interaction with our reanimated ELIZA.

Once the script has been edited it is useful to save it. Giving no input (just pressing ENTER at the INPUT prompt) will cause ELIZA to ask for a pseudo-tape number to write its rules to. Entering 0 (or giving no input at all) lead it to print the script to the console, and entering a number, say 130, will write to disk file: \texttt{.TAPE.130}. The output is a well-formed S-expressions, but with different spacing than we usually see in external scripts that have been edited by hand, or reformatted (sometimes incorrectly!) for publication. This can also be observed in the Appendix.

Although this `teaching' mode was completely implemented, and appears to work, it is only briefly mentioned in the the 1966 paper, and is nowhere else described. Weizenbaum appears to have abandoned this internal teaching method, given the availability of text editors in CTSS. Indeed, elsewhere in the 1966 paper he describes invoking ED from ELIZA by using the EDIT command. This does not exist in the found code.

\section{ELIZA's Final Analysis}\label{conc}

While simple by today's standards, ELIZA was groundbreaking for its time, bringing to life the concept -- mere science fiction to that point –- of conversation with a computer, and igniting a fire (for better or worse) under AI. ELIZA had a significant, if unintended and somewhat indirect, impact on AI. It was the first program to directly embody Turing's test and brought into reality what was to that point (and to a large extent, thereafter as well, until perhaps even a couple of years ago) the idea that you could have an actual conversation with a computer. It showed, in part, in its overt stupidity that you may not have to actually solve intelligence to obtain artificial intelligence -- up to that point, and up to the spectacular and unexpected apparent intelligence of LLMs -- people were trying to model how people thought and code those processes into computers. There were almost certainly other indirect impacts from ELIZA; it is embedded in the AI psyche.

While we can't put ourselves in the shoes of someone living in 1966, it is fascinating to experience a conversation with the original ELIZA, even if it breaks on occasion. If the code we discovered didn't work at all it would be very disappointing and we'd be trying to figure out why. 

We have have struggled with whether to repair the problem with inputting numbers. Our feeling is that running the original code feels good and authentic. Finding bugs in it only adds to the authenticity. Any changes or bug fixes we make detract from that, except minor interventions necessary to get the code running so people can run at all. Put in more academic terms, it is important that programs discovered in archaeological research of this sort retain as much of their exact nature as possible. It is true that we needed to make repairs at the margin in order to get the discovered code running, but as was mentioned, the running code is 96\% exactly what was discovered, and what was changed is clearly documented, but there were no changes made to the core code as would be required to repair the number problem. 

Of course, we continue to search various archives for later, or even earlier, versions of Eliza, and we would appreciate it if any readers of this paper who may have access to archives that may potentially contain versions of Eliza would make contact with us. 

\section{DIY ELIZA}

Here is a video of the reanimated ELIZA holding the conversation from the 1966 CACM paper: \cite{DemoVid}. You can download and build CTSS and ELIZA to run on your own computer here: \cite{ELIZACTSSRepo}. This has been tested on various Linux and MacOS versions, but we've noticed some issues with different versions, so your mileage may vary. If you get it working on your machine and find that you have to change something, let us know. (Or fork the repo and send us a pull request!) Report discoveries, issues (or even better, fixes!) to rupert@timereshared.com.

\pagebreak

\section*{Acknowledgments}

This work could not have happened without the support and contribution of Team ELIZA, a research group looking into the history of the original ELIZA. Team ELIZA is David M. Berry, Sarah Ciston, Anthony Hay, Mark C. Marino, Peter Millican, Arthur Schwarz, Jeff Shrager, and Peggy Weil. Thanks to Dave Pitts for his amazing 7094 emulator and CTSS kit, to Jerry Saltzer and Tom Van Vleck for contributing their deep knowledge of CTSS and its history, to the MIT archivists esp. Myles Crowley and Allison Schmitt, and to Pm Weizenbaum and the Weizenbaum estate for permission to open source the code discovered in Prof. Weizenbaum's archives.

\section*{Author Contributions}

JS organized the team and discovered the original code in Joseph Weizenbaum's archives. DMB recognized the opportunity to bring the original ELIZA up on an emulator, made the initial connection to RL, and joined JS in a second spelunking tour of the archives. RL did all of the hands-on work involved in bringing up the emulator and compiling and debugging ELIZA. His efforts were significantly supported by the extensive background work and knowledge, and additional research by AS and AH, who had spent several years, studying the code and had extensive prior knowledge of ELIZA and SLIP. AH had previously written the most accurate known ELIZA clone, and AS has written several implementations of SLIP, including the GNU implementation, GNUSLIP. JS composed this paper (largely from email discussions among the authors). RL, JS, and AH have tested the code. All of them have read, improved, and approved this manuscript.

\pagebreak

\printbibliography

\pagebreak

\section*{Appendix: A complete example of teaching ELIZA new rules}
[Comments are enclosed in square brackets.]

\begin{verbatim}
% telnet 0 7094   [CTSS is listening to local port 7094.]                                     
Trying 0.0.0.0...
Connected to 0.
Escape character is '^]'.
s709 2.4.1 COMM tty0 (KSR-37) from 127.0.0.1

MIT8C0: 1 USER AT 01/08/14 1915.8, MAX = 30
READY.

login eliza
W 1915.9
Password
 YOU HAVE      10L SAVED
 M1416    10 LOGGED IN  01/08/14 1915.9 FROM 700000
 LAST LOGOUT WAS  12/24/14  048.6 FROM 700000
 HOME FILE DIRECTORY IS M1416 ELIZA

THIS IS A RECONSTRUCTED CTSS SYSTEM.
IT IS A DEBUG AND NOT FULLY FUNCTIONAL VERSION.

 CTSS BEING USED IS: MIT8C0
R .016+.000

r eliza
W 1921.6
EXECUTION.
WHICH SCRIPT DO YOU WISH TO PLAY
200
HOW DO YOU DO . PLEASE TELL ME YOUR PROBLEM
INPUT
i want a pony

WHAT WOULD IT MEAN TO YOU IF YOU GOT A PONY
INPUT
* hi              [Adding a new rule via * 'fast' mode]
((hi eliza) (how do you do))
((hi 0) (hello from 2))

INPUT
hi eliza

HOW DO YOU DO     [This hits the first transformation]
INPUT
hi there you fun program you
                  [This hits the second transformation]
HELLO FROM THERE I FUN PROGRAM I
INPUT
+                 [Entering 'teaching' mode]

PLEASE INSTRUCT ME
INPUT
type hi
                   [Display our new rule]
HI

HI ELIZA
  1
HOW DO YOU DO

HI 000000
  1
HELLO FROM 000002
PLEASE INSTRUCT ME
INPUT              [Change one of the tranformation rules in 'hi']
subst hi (hello from 2) (goodbye 2)

PLEASE INSTRUCT ME
INPUT
type hi            [Let's check that it's actually changed]

HI

HI ELIZA
  1
HOW DO YOU DO

HI 000000
  1
GOODBYE 000002      [Looks good]
PLEASE INSTRUCT ME
INPUT
start               [Back to conversation mode to test our new rules]

INPUT
hi to all my friends
                    [Oops! 'my' has higher priority (2)]
YOUR FRIENDS
INPUT
hi to the cats of the world

GOODBYE TO THE CATS OF THE WORLD
INPUT
+
rank hi 9           [Let's rank 'hi' above 'my']

PLEASE INSTRUCT ME
INPUT
type hi
                    [Check that the re-rank worked]
HI 000009

HI ELIZA
  1
HOW DO YOU DO

HI 000000
  1
GOODBYE 000002
PLEASE INSTRUCT ME
INPUT
start

INPUT
hi to all my friends

GOODBYE TO ALL YOUR FRIENDS   [Looks good!]
INPUT
+
                    [Let's change one of the given rules]                     
PLEASE INSTRUCT ME
INPUT
type your

YOUR = MY

000000 MY 000000
WHY ARE YOU CONCERNED OVER MY 000003
WHAT ABOUT YOUR OWN 000003
ARE YOU WORRIED ABOUT SOMEONE ELSES 000003
REALLY , MY 000003
PLEASE INSTRUCT ME
INPUT
start
my foot itches

INPUT
my foot itches

WHY DO YOU SAY YOUR FOOT ITCHES
INPUT
your foot itches

WHY ARE YOU CONCERNED OVER MY FOOT ITCHES
INPUT
+                    [Changing one of the rules for YOUR/MY]                     
subst your (WHY ARE YOU CONCERNED OVER MY 3) (worry about your own 3)

PLEASE INSTRUCT ME
INPUT
type your            [Did it take?]                     


YOUR = MY

000000 MY 000000
  1
WORRY ABOUT YOUR OWN 000003
WHAT ABOUT YOUR OWN 000003
ARE YOU WORRIED ABOUT SOMEONE ELSES 000003
REALLY , MY 000003
PLEASE INSTRUCT ME
INPUT
start                [Let's test it!]

INPUT
your spoon is too big
                     [We may have to roll through all the transformations]
WHAT ABOUT YOUR OWN SPOON IS TOO BIG
INPUT
your spoon is too big

ARE YOU WORRIED ABOUT SOMEONE ELSES SPOON IS TOO BIG
INPUT
your spoon is too big

REALLY , MY SPOON IS TOO BIG
INPUT
your spoon is too big
    
WORRY ABOUT YOUR OWN SPOON IS TOO BIG     [Tada!]
INPUT
[User enters a blank line (double ENTER) causing ELIZA to dump the script.]
WHAT IS TO BE THE NUMBER OF THE NEW SCRIPT
[The user does not enter a number, so the script is dumped to the teletype.]
[For brevity's sake we only show the affected entries.]
(   HI          000009        (     (   HI          ELIZA         )
000001        (   HOW         DO          YOU         DO            )
  )     (     (   HI          000000        )   000001        (   GOODBY
E           000002        )     )     )
[...]
(   YOUR        =           MY            (     (   000000      MY
      000000        )   000001        (   WORRY       ABOUT       YOUR
      OWN         000003        )     (   WHAT        ABOUT       YOUR
      OWN         000003        )     (   ARE         YOU         WORRIE
D           ABOUT       SOMEONE           ELSES       000003        )
  (   REALLY      ,           MY          000003        )     )     )
[...]
  EXIT CALLED. PM MAY BE TAKEN.
R 1.250+.016
\end{verbatim}

\end{document}